\pdfoutput=1
\documentclass[11pt]{article}

\usepackage[]{naacl2021}

\usepackage{times}
\usepackage{latexsym}
\usepackage{graphicx}
\usepackage[T1]{fontenc}
\usepackage[utf8]{inputenc}

\usepackage{microtype}

\title{¡Qué maravilla! Multimodal Sarcasm Detection in Spanish: a Dataset and a Baseline}

\author{Khalid Alnajjar \\
  Department of Digital Humanities \\
  University of Helsinki \\
  \texttt{khalid.alnajjar@helsinki.fi} \\\And
  Mika Hämäläinen \\
  Department of Digital Humanities \\
  University of Helsinki \\
  \texttt{mika.hamalainen@helsinki.fi} \\}

\begin{document}
\maketitle
\begin{abstract}
We construct the first ever multimodal sarcasm dataset for Spanish. The audiovisual dataset consists of sarcasm annotated text that is aligned with video and audio. The dataset represents two varieties of Spanish, a Latin American variety and a Peninsular Spanish variety, which ensures a wider dialectal coverage for this global language. We present several models for sarcasm detection that will serve as baselines in the future research. Our results show that results with text only (89\%) are worse than when combining text with audio (91.9\%). Finally, the best results are obtained when combining all the modalities: text, audio and video (93.1\%).
\end{abstract}

\section{Introduction}

Figurative language is one of the most difficult forms of natural language to model computationally and there have been several studies in the past focusing on its subcategories such as metaphor interpretation \cite{xiao2016meta4meaning,hamalainen-alnajjar-2019-lets}, humor generation \cite{hamalainen2019modelling} and analyzing idioms \cite{flor2018catching}. Sarcasm is one of the extreme forms of figurative language, where the meaning of an utterance has little to do with the surface meaning (see \citealt{kreuz1989sarcastic}). 

Understanding sarcasm is difficult even for us humans as it requires certain mental capacities such as a theory of mind (see \citealt{ZHU2020110035}) and it is very dependent on the context and speaker who is being sarcastic. There are also very different view to sarcasm in the literature, for example, according to \citet{kumon1995another} sarcasm requires an allusion to a failed expectation and pragmatic insincerity (see \citealt{grice1975logic}) to be present in the same time. However, \citet{utsumi1996unified} highlights that these two preconditions are not enough, as sarcasm needs an ironic context to take place.

\citet{haverkate1990speech} argues that, in the context of sarcasm, the meaning difference can either be the complete opposite of the semantic meaning of a sentence or somewhat different as seen in the lexical opposition of the words and the intended meaning. The fact that there are several different theoretical ways of understanding sarcasm, highlights the complexity of the phenomenon.

In this paper, we present an audio aligned dataset for sarcasm detection in Spanish. The dataset containing text and video timestamps has been released openly on Zenodo\footnote{Open access version of the data (contains text only) https://zenodo.org/record/4701383}. An access to the dataset with the video clips\footnote{Access by request version of the data (videos and text) https://zenodo.org/record/4707913} can be granted upon request for academic use only. In addition, we will present a baseline model for this dataset to conduct multimodal sarcasm detection in Spanish.

\begin{table*}[ht]
\centering
\small
\begin{tabular}{|l|l|l|l|}
\hline
Speaker & Utterance                                                                                                                       & Translation                                                                                                                    & Sarcasm \\ \hline
Archer  & No, Lana, para nada                                                                                                             & No, Lana, not at all                                                                                                           & true    \\ \hline
Stan    & \begin{tabular}[c]{@{}l@{}}Lo siento chicos, mi papá dice que está muy \\ ocupado con los Broncos, no tiene tiempo\end{tabular} & \begin{tabular}[c]{@{}l@{}}I am sorry guys, my dad says he is very \\ busy with the Broncos, he doesn't have time\end{tabular} & false   \\ \hline
Lana    & Decías algo acerca de un plan                                                                                                   & You said something about a plan                                                                                                & true    \\ \hline
\end{tabular}
\caption{Example sentences from the dataset.}
\label{tab:data-example}
\end{table*}

\section{Related work}

In this section, we will present some of the recent related work on sarcasm detection. There has been some work also on sarcasm generation \cite{chakrabarty-etal-2020-r} and interpretation \cite{peled-reichart-2017-sarcasm}, but they are rather different as tasks and we will not discuss them in detail.

\citet{badlani-etal-2019-ensemble} show an approach for sarcasm detection in online reivews. They train a CNN (convolutional neural network) based model on separate feature embeddings for sarcasm, humor, sentiment and hate speech. Similarly, \citet{babanejad-etal-2020-affective} also detect sarcasm in text. They combine an LSTM (long short-term memory) model with BERT. \citet{dubey-etal-2019-numbers} also work on text only by detecting sarcastic numbers in tweets. They experiment with rules, SVMs (support vector machines) and CNNs.

\citet{cai-etal-2019-multi} use an LSTM model to detect sarcasm in tweets. Their approach is multimodal in the sense that it takes text and images into account, but it does not deal with audio and video like our approach.

\citet{castro2019towards} present a multimodal sarcasm dataset in English. The dataset consists of annotated videos from TV sitcoms such as Friends and the Big Bang Theory, apart from being in English instead of Spanish, the main difference is that our dataset consists of animated cartoons instead of TV shows played by real people. Another big difference is in the data collection as they opted for querying sarcastic video clips, where as the data we work with represents full episodes. \citet{chauhan-etal-2020-sentiment} use this data and present a multimodal sarcasm detection framework based on a Bi-GRU model.

Many of the related work has been focusing on text only. Research on multimodal approaches has been carried out only for English data, not unlike the textual approaches.

\section{Dataset}

We base our work on the sarcasm annotated dataset from the MA thesis of the second author of this paper \citet{sarcasmo}\footnote{Available on https://www.kaggle.com/mikahama/the-best-sarcasm-annotated-dataset-in-spanish}. This dataset is based on two episodes of South Park with voice-overs in Latin-American Spanish and two episodes of Archer with voice-overs in Spanish of Spain. The dataset has the speaker, their utterance and sarcasm annotations for each utterance in all of the episodes. However, the released data has been released shuffled for copyright reasons and it contains text only. Unlike the recent multimodal dataset for English \cite{castro2019towards}, this data is expert annotated according to several different theories on sarcasm. 

Annotation based on theories is important in order to establish a concrete meaning for sarcasm and to avoid the same mistakes as \citet{castro2019towards} had. In their paper, they report that the most sarcastic character in The Big Bang Theory is Sheldon, however this cannot be true as one of the main characteristics of Sheldon is that he does not understand sarcasm. Therefore, their annotations ignore the fundamentally important characteristic of sarcasm, which is speaker intent, and rather they consider sarcasm purely based on subjective intuition. 

In order to produce a multimodal dataset out of the existing one, we locate the corresponding videos for the annotations and manually align them with the video. We use our own in-house tool JustAnnotate for this task\footnote{https://mikakalevi.com/downloads/JustAnnotate.exe}. This was a time consuming task, but as a result we ended up with a high-quality dataset with audio, video and text aligned. While aligning the dataset, we found several errors in the original transcriptions that we fixed. We did not alter the sarcasm annotations. In addition to the alignment, we introduced scene annotations. An episode of a TV show consists of many different scenes, and sarcasm is typically highly contextual, we indicate in the data which utterances belong to the same scene to better capture the context of each utterance.

Table \ref{tab:data-example} shows an example of the dataset. The English translation is provided for convenience, but it is not included in the dataset itself. Each line is aligned with audio and video. As we can see from these examples, sarcasm in the dataset is very contextually dependent as the sarcastic sentences presented in the table might equally well be sincere remarks if uttered by someone else or in a different context.

\begin{figure}[h]
  \centering
  \includegraphics[width=\linewidth]{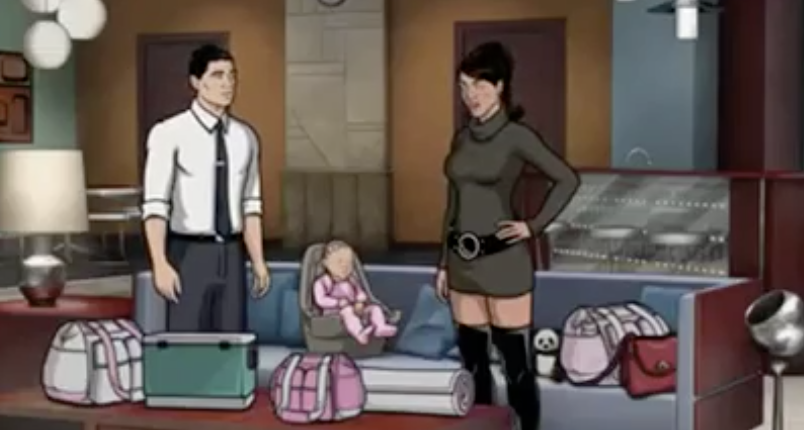}
  \caption{Archer uttering a sarcastic sentence that goes against the common sense}
  \label{example_archer}
\end{figure}

Figure \ref{example_archer} shows an example of a scene in the corpus. In this particular scene, Archer asks sarcastically \textit{¿Dónde se compra la leche materna?} (Where does one buy breast milk?). This is an example of sarcasm in the corpus where sarcasm violates common sense. Depending on the speaker, the utterance might be sarcastic or the speaker might lack knowledge on the topic.

\begin{figure}[h]
  \centering
  \includegraphics[width=\linewidth]{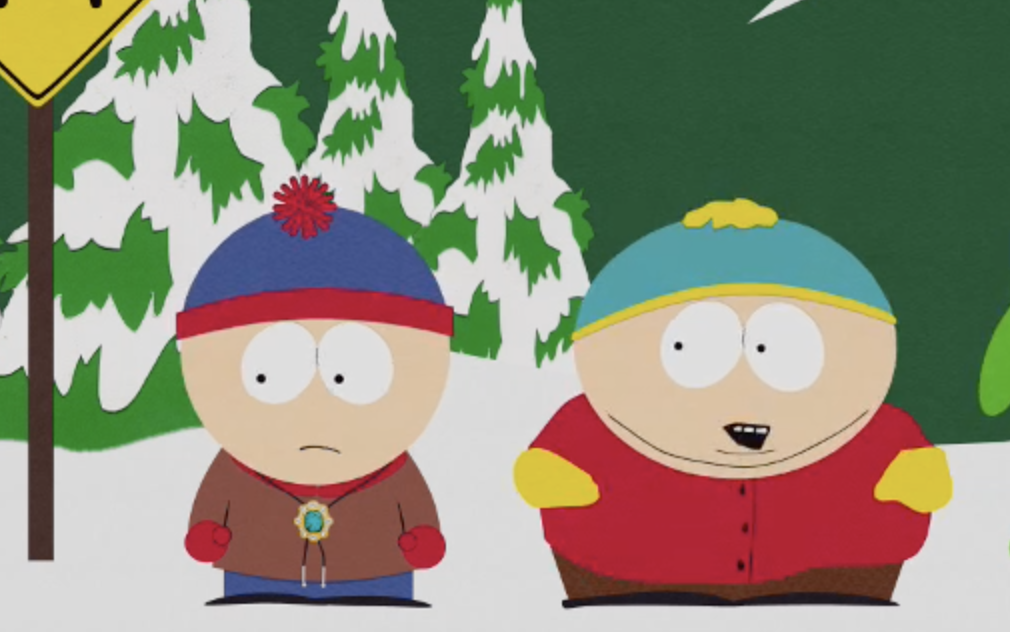}
  \caption{Cartman uttering a sarcastic sentence that can be resolved only by visual cues.}
  \label{example_south}
\end{figure}

In Figure \ref{example_south} Cartman comments on the neckpiece of Stan by saying \textit{Esas corbatas están de moda, tiene suerte de tenerla} (Those neckpieces are fashionable, you are lucky to have one). This is an example of a very different type of sarcasm that cannot be detected just by having common knowledge about the world. In order to understand the sarcastic intent, a system would need to have an access to the video as well to detect the unfashionable neckpiece and the disappointed facial expression of Stan.

\section{Method}

In this section, we present our method for detecting sarcasm in the multimodal dataset. We experiment with text only, text and audio and all modalities. All models are trained by using the same random train (80\%) and test (20\%) splits. For the neural model, 10\% of the training split is used for validation.

\subsection{Text only}

For the text only model, we experiment with two models. In the first one, we use an off the shelf OpenNMT \cite{opennmt} model. We train the model using a bi-directional long short-term memory (LSTM) based model \cite{hochreiter1997long} with the default settings except for the encoder where we use a BRNN (bi-directional recurrent neural network) \cite{schuster1997bidirectional} instead of the default RNN (recurrent neural network). We use the default of two layers for both the encoder and the decoder and the default attention model, which is the general global attention presented by \citet{luong2015effective}. The model is trained for the default 100,000 steps.

The second model is a Support Vector Machine (SVM) \cite{10.1162/089976600300015565}, due to its efficiency when dealing with a high dimensional space and ability to train a model with small data. We use the SVM implementation provided in Scikit-learn~\cite{scikit-learn}. Following the work of~\citet{castro2019towards}, we use an RBF kernel and a scaled gamma. The regularization parameter $C$ is set for 1000. This setup is followed in all of our SVM models.

Regarding the textual features of the SVM, we make use of GloVe~\cite{pennington-etal-2014-glove} embeddings\footnote{\url{https://github.com/dccuchile/spanish-word-embeddings}} trained on the Spanish Billion Words Corpus \cite{cardellinoSBWCE} and ELMo~\cite{Peters:2018} embeddings provided by~\cite{che-EtAl:2018:K18-2}. Each textual instance is tokenized using TokTok\footnote{\url{https://github.com/jonsafari/tok-tok}}, and then a sentence-level vector is constructed by computing the centroid (i.e., average vector) of all tokens, for each word embeddings type. In the case of ELMo, the vector of each token is the average of the last three layers of the neural network. The input to the SVM model is the concatenation of the two types of sentence embeddings.

\subsection{Text and audio}

This model is an SVM based model that extends the textual SVM model with audio features. We do not extend the OpenNMT model with audio features as the library does not provide us with audio and video inputters.

For all the audio, we set their sample size into 22 kHz to convert the data into a manageable and consistent size. Thereafter, we extract different audio features using librosa~\cite{brian_mcfee_2020_3955228}. These features include short-time Fourier transform \cite{10.5555/42739.42745}, mel-frequency cepstral coefficients \cite{stevens1937scale}, chroma, Tonnetz \cite{harte2006detecting}, zero-crossing rate, spectral centroid and bandwidth, and pitches. In total, 13 features\footnote{We used the following methods from librosa: \textit{stft, mfcc, chroma\_stft,  spectral\_centroid, spectral\_bandwidth, spectral\_rolloff, zero\_crossing\_rate, piptrack, onset\_strength, melspectrogram, spectral\_contrast, tonnetz} and \textit{harmonic}} were extracted. By combining all these features, we get the audio vector.

\subsection{All modalities}

For videos, instead of trying to represent an entire video as a vector like some of the existing approaches \cite{hu2016video2vec} to video processing, we extract 3 frames for each video corresponding to an utterance. We extract the frames by dividing the frames of a video clip into three evenly sized chunks and taking the first frame of each chunk. The key motivation behind this is that we are working with animation, where most of the frames are static and changes in between frames are not big. Therefore representing the entire video clip is not important and it would only increase the complexity of the system.

We extract visual features from each of the three frames extracted using a pre-trained ResNet-152 model~\cite{7780459}. Features are taken from the last layer in the network, and the overall video vector is the sequence of the three feature embeddings, in the same order. All the vectors described above (i.e., textual, audio and visual vectors) are passed as input to the all-modalities SVM model.

\section{Results}

In this section, we report the accuracy of predictions by the neural model and the three SVM models that are based on 1) text only, 2) text and audio, and 3) text, audio and video. The results can be seen in Table \ref{tab:results}.

%OpenNMT 0.875

\begin{table}[h]
\centering
\begin{tabular}{|l|c|}
\hline
\textbf{Input}       & \textbf{Accuracy} \\ \hline
\multicolumn{2}{|c|}{Neural Model}       \\ \hline
Text                 & 87.5\%             \\ \hline
\multicolumn{2}{|c|}{SVM}                \\ \hline
Text                 & 89.0\%             \\ \hline
Text + Audio         & 91.9\%             \\ \hline
Text + Audio + Video & \textbf{93.1\%}    \\ \hline
\end{tabular}
\caption{Accuracies of the predictions by all models for the sarcasm detection task.}
\label{tab:results}
\end{table}

As we can see in the results, having more modalities in the training improved the results. The audio features were able to capture more features important for sarcasm than pure text. Having all the three modalities at the same time gave the best results, with a 4.1\% gain in the accuracy from the text-based model. The neural model reached to the lowest accuracy, most likely due to the fact that it was not trained with pretrained embeddings, a source of information that was available to the SVM models.

\subsection{Error analysis}

When we look at the predictions by the model best model (text + audio + video), we can see that the sarcasm detection is not at all an easy task.

An interesting example of a correctly predicted sarcastic utterance is \textit{Lucen bien muchachos. ¡A patear culos!} (You look great, guys. Let's kick some ass!). This is an example of a visually interpretable sarcasm where the kids the sentence was uttered to looked all ridiculous. This would seem, at first, to highlight that the model has learned something important based on the visual features. However, we can see that this is not at all the case as the model predicts incorrectly the following sarcastic utterance: \textit{Sí Stan, es lo que quiere la gente. No te preocupes, luces genial.} (Yes Stan, that is what the people want. Don't worry, you look great.) The context is similar to the one where the model predicted the sarcasm correctly, which means that the visual features are not representative enough for the model to correctly annotate detect this sarcastic utterance.

Interestingly, the model predicted \textit{Sí amigo, es una réplica de la corbata del Rey Enrique V} (Yes friend, it is a replica of the neckpiece of the King Henry V) as sarcastic while in fact the uttrance was not sarcastic. This utterance refers to the same neckpiece as seen in Figure \ref{example_south}. The neckpiece appeared frequently in sarcastic contexts, so the model overgeneralized that anything said about the neckpiece must be sarcastic.

\section{Conclusions}

We have presented the first multimodal dataset for detecting sarcasm in Spanish. The dataset has been released  on Zenodo. Our initial results serve as a baseline for any future work on sarcasm detection on this dataset.

Based on the results, it is clear that multimodality aids in detecting sarcasm as more contextual information is exposed to the model. Despite the improvements when considering multiple modalities, sarcasm detection is a very difficult task to model as it demands a global understanding of the world and the specific context the sarcastic utterance is in, as discussed in our error analysis. Even though the overall accuracy is high, it is clear the model makes errors that indicate that it has learned the data, but not the phenomenon.
%, which is presumably the case for most of the NLP classification models out there.

\bibliography{anthology,custom}
\bibliographystyle{acl_natbib}

\end{document}